\documentclass[twoside,11pt]{article}

\usepackage{blindtext}

\usepackage[preprint]{jmlr2e}
\usepackage{seqsplit}

\usepackage[dvipsnames]{xcolor}
\definecolor{Gray}{gray}{0.9}
\definecolor{mygreen}{rgb}{0.0, 0.5, 0.0}
\definecolor{myred}{rgb}{0.8, 0.25, 0.33}
\definecolor{myblue}{rgb}{0.19, 0.55, 0.91}
\definecolor{uclablue}{rgb}{0.15, 0.45, 0.68}
\definecolor{boxgreen}{rgb}{0.02, 0.66, 0.02}
\definecolor{boxred}{rgb}{0.66, 0.1, 0.1}
\definecolor{boxblue}{rgb}{0.01, 0.01, 0.73}
\definecolor{mygray}{gray}{0.4}
\usepackage{notation}
\usepackage{cuted}
\usepackage{hyperref}
\usepackage{marvosym}
\usepackage{graphicx}
\usepackage{amssymb}
\usepackage{url}

\usepackage{microtype}
\usepackage{subfigure}
\usepackage{booktabs}

\usepackage{amsmath}
\usepackage{mathtools}
\usepackage{amsthm}

\usepackage{listings}
\usepackage{etoc}
\usepackage{algorithm}
\usepackage{algorithmic}
\usepackage{lipsum}
\usepackage{pifont}
\usepackage{wrapfig}
\usepackage{colortbl}
\usepackage{acronym}
\usepackage{multicol}
\usepackage{multirow}
\usepackage{makecell}
\usepackage{enumitem}
\usepackage{tabularx}
\usepackage{tabularray}
\UseTblrLibrary{booktabs}
\usepackage{arydshln}
\usepackage{colortbl}
\usepackage{array}
\usepackage[skip=3pt,font=small]{caption}
\usepackage{cleveref}
\usepackage{xspace}
\usepackage{mdframed}

\usepackage[export]{adjustbox}

\newcolumntype{Y}{>{\arraybackslash}X}

\usepackage[utf8]{inputenc} 
\usepackage[T1]{fontenc}    
\usepackage{hyperref}       
\usepackage{amsfonts}       
\usepackage{nicefrac}       

\usepackage{mathtools}
\usepackage{amssymb}
\usepackage{amsthm}

\makeatletter
\DeclareRobustCommand\onedot{\futurelet\@let@token\@onedot}
\def\@onedot{\ifx\@let@token.\else.\null\fi\xspace}

\makeatother

\acrodef{llms}[LLMs]{Large Language Models}
\acrodef{mlms}[MLMs]{Multimodal Language Models}

\hypersetup{ hidelinks }

\newcommand{\method}{\textsc{MineStudio}\xspace}

\usepackage{lastpage}
\jmlrheading{23}{2025}{1-\pageref{LastPage}}{1/21; Revised 5/22}{9/22}{21-0000}{Cai, Mu, He, Zhang, Zheng, Liu and Liang}

\ShortHeadings{\method: A Streamlined Package for Minecraft AI Agent Development }{Cai, Mu, He, Zhang, Zheng, Liu and Liang}
\firstpageno{1}

\begin{document}

\title{
\method: A Streamlined Package for \\Minecraft AI Agent Development 
}

\renewcommand{\today}{} 





\author{\name Shaofei~Cai$^{1,*}$ \email caishaofei@stu.pku.edu.cn \\
       \name Zhancun~Mu$^{1,*}$ \email muzhancun@stu.pku.edu.cn \\
       \name Kaichen~He$^{1}$ \email hekaichen@stu.pku.edu.cn \\
       \name Bowei~Zhang$^{1}$ \email zhangbowe@stu.pku.edu.cn \\
       \name Xinyue~Zheng$^{1}$ \email zhengxinyue@bigai.ai \\
       \name Anji~Liu$^{2}$ \email liuanji@cs.ucla.edu \\
       \name Yitao~Liang$^{1,\dagger}$ \email yitaol@pku.edu.cn \\
       \addr
        $^{1}$ Institute for Artificial Intelligence, Peking University \\
        $^{2}$ University of California, Los Angeles \\
        $^*$ Equal contribution\quad $\dagger$ Corresponding author
}

\editor{Submitted to JMLR OSS}

\maketitle

\begin{abstract}%
Minecraft's complexity and diversity as an open world make it a perfect environment to test if agents can learn, adapt, and tackle a variety of unscripted tasks. However, the development and validation of novel agents in this setting continue to face significant engineering challenges. 
This paper presents \method, an open-source software package designed to streamline the development of autonomous agents in Minecraft. \method represents the first comprehensive integration of seven critical engineering components: simulator, data, model, offline pre-training, online fine-tuning, inference, and benchmark, thereby allowing users to concentrate their efforts on algorithm innovation. We provide a user-friendly API design accompanied by comprehensive documentation and tutorials. Our project is released at \href{https://github.com/CraftJarvis/MineStudio}{https://github.com/CraftJarvis/MineStudio}. 

\end{abstract}


\section{Introduction}
\label{sec:intro}

The development of highly capable and adaptable artificial intelligence (AI) agents remains a significant frontier in AI research. While impressive successes have been demonstrated across various domains, such as large language models like ChatGPT~\citep{gpt3} and advancements in protein folding \citep{alphafold}, transferring these achievements to building agents that interact effectively across a wide range of challenges in complex, dynamic environments is a crucial next step \citep{vpt, minedojo, voyager}. In this context, open-world environments, particularly Minecraft \citep{johnson2016malmo, minerl}, have emerged as unparalleled testbeds.
In this context, open-world environments, particularly Minecraft~\citep{johnson2016malmo, minerl}, have emerged as unparalleled testbed. 
Unlike Procgen~\citep{cobbe2019procgen} and Habitat~\citep{habitat}, Minecraft's procedural generation, infinite scale, and diverse interaction possibilities offer a rich, unscripted setting that presents agents with (i) an unprecedented variety of tasks and (ii) notably more complex and long-horizon challenges \citep{deps, jarvis-1}. Here, agents must learn to perceive, plan, and act over extended durations, tackling a myriad of emergent tasks from basic resource gathering to elaborate construction. 
This intrinsic complexity makes it an ideal platform to push the boundaries of current AI capabilities. 

However, leveraging Minecraft for cutting-edge AI research is far from trivial. Researchers frequently face substantial engineering challenges in setting up robust simulation environments~\citep{minerl}, managing large-scale data~\citep{vpt}, developing complex offline pre-training and online fine-tuning pipelines, and consistently evaluating agent performance in such an open-ended setting~\citep{mcu}. These overheads often divert valuable time and resources away from the core algorithmic innovation that drives progress in AI. The fragmentation of essential components across different tools and platforms further exacerbates this issue, creating significant barriers to entry and hindering reproducible research. 

To address these bottlenecks, we present \method, an open-source software package meticulously designed to streamline the entire lifecycle of autonomous agent development in Minecraft. \method stands out by providing the first comprehensive integration of seven core engineering components: easily customizable simulator, trajectory data management, model templates, offline pre-training pipelines, online fine-tuning mechanisms, efficient inference, and standardized benchmarking. By consolidating these disparate elements into a unified, user-friendly framework, \method empowers researchers to focus their efforts on algorithmic innovation, accelerating the development of next-generation intelligent agents. 

\section{\method}
\label{sec:minestudio}

\begin{figure*}[t]
\begin{center}
\includegraphics[width=0.95\linewidth]{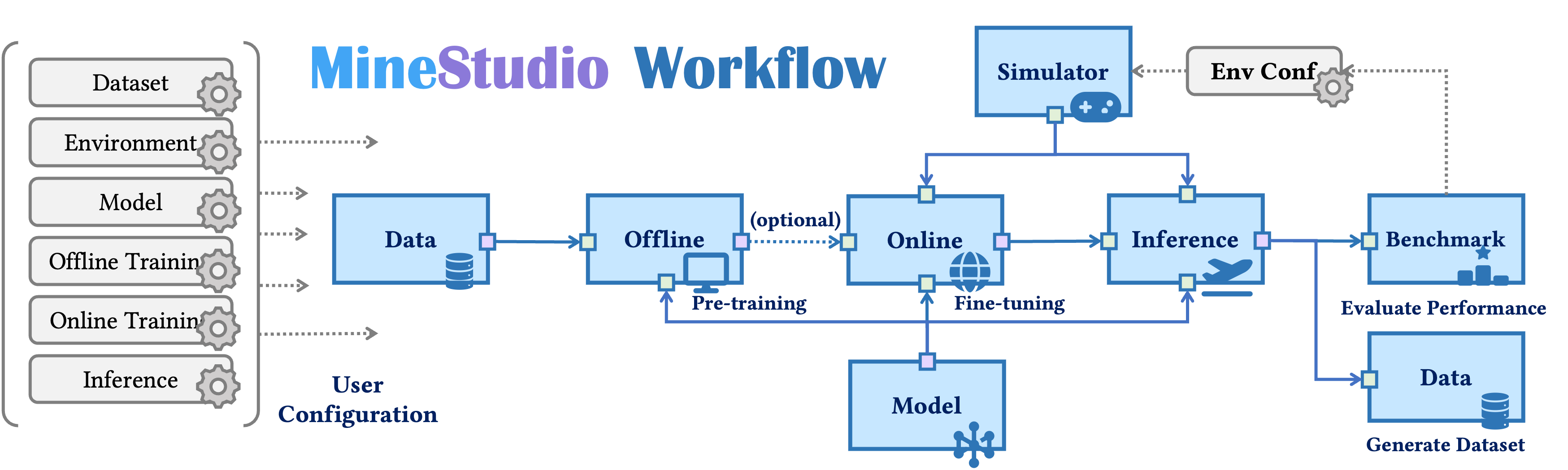}
\end{center}
\vspace{-0.8em}
\caption{
\method enables users to address classic requirements such as offline pertaining and online fine-tuning with minimal coding effort. Users only need to configure the model component with a small amount of PyTorch code. Each module in the workflow is fully customizable, allowing users to configure settings or extend functionality through inheritance and overrides as needed. 
}
\vspace{-0.5em}
\label{fig:workflow}
\end{figure*}

\method is an efficient, end-to-end toolkit designed to streamline the development of Minecraft AI agents, encompassing the entire workflow from data preparation and training to evaluation. Its flexible interface not only provides a wealth of out-of-the-box features but also empowers users to tailor the toolkit to their specific research needs. Figure \ref{fig:workflow} illustrates the overall pipeline, with subsequent paragraphs detailing each module. 

\noindent \paragraph{Simulator.} This component implements a hook-based Minecraft simulator wrapper that offers a high degree of customization. By subclassing \texttt{MinecraftCallback}, users can tailor the environment in various ways, including (but not limited to) monitoring rendering frame rate, issuing commands, modifying the terrain, logging episodes, generating complex reward functions, and overriding observations. To minimize setup effort, we include a suite of commonly used callbacks. We also incorporate several rendering optimizations to speed up evaluation, data collection, and reinforcement learning. 

\noindent \paragraph{Data.} We provide an efficient and flexible structure for managing offline trajectory data. 
Conventional trajectory storage methods either save frames as individual image files or as complete videos. The former consumes excessive resources, while the latter doesn't support fast reading. We segment trajectories into short clips and store them independently in \href{https://lmdb.readthedocs.io/}{LMDB} files, preserving temporal alignment. This design balances storage efficiency with fast video decoding, enabling quick retrieval of trajectory segments by semantic label. 
To support models requiring long-term memory, we include a distributed batch sampler capable of streaming long sequences. We also offer data conversion scripts to help users integrate their own trajectory datasets, along with visualization tools for debugging.
The framework includes the largest available Contractor Dataset \citep{vpt} in the Minecraft domain, complemented by frame-level semantic segmentation data generated by \cite{rocket1}. 

\noindent \paragraph{Model.} This component provides a unified template for policy networks along with action and value heads designed for Minecraft, allowing users to focus on the architecture design. We have pre-integrated models such as VPTs~\citep{vpt}, STEVE-1~\citep{steve1}, GROOT-1~\citep{groot1}, ROCKETs~\citep{rocket1, rocket2}. The standardized interface functions ensure that policies can be seamlessly integrated with other \method modules, enabling efficient training and inference as well as fair comparisons. 

\noindent \paragraph{Offline Pre-Training.} This component builds upon the \href{https://lightning.ai/}{PyTorch Lightning} \citep{pytorch_lightning} framework to deliver an enhanced Trainer module. We incorporate dedicated mechanisms for handling policy memory (e.g., TransformerXL \citep{transformerxl}) and integrate seamlessly with the distributed batch sampler in the \method data module, thereby enabling training on ultra-long trajectories. In addition, we provide a proven set of hyperparameter configurations—covering aspects such as warmup steps, optimizers, and learning rates—while also allowing flexible customization through objective callbacks. 

\noindent \paragraph{Online Fine-Tuning.} This component implements the KL-constrained Proximal Policy Optimization algorithm introduced in \cite{vpt}. The code is designed for seamless integration with our model and simulator. It has been optimized for memory-based policies, enabling efficient storage and management of memory states over long episodes. It is also designed to handle the inherent instability of the Minecraft environment.
Additionally, we provide a set of hyperparameters that have been empirically validated to ensure high training efficiency and strong performance across diverse tasks. We hope this resource serves as an accessible starting point and lowers the barrier to RL research in Minecraft.

\noindent \paragraph{Inference.} We provide a \href{https://docs.ray.io/}{Ray}-based \citep{moritz2018ray} inference framework to support distributed inference, consisting of three parts: \texttt{generator}, \texttt{filter}, and \texttt{recorder}, forming an asynchronous inference pipeline for easily evaluating the performance of different agents. The \texttt{generator} part, equipped with an agent creator and an environment creator, generates trajectories in batches. Each produced trajectory is immediately passed to a \texttt{filter} for post-processing and then summarized and stored by the \texttt{recorder}. By customizing the \texttt{filter} and \texttt{recorder}, users can effortlessly conduct comprehensive evaluations of policy checkpoints. Furthermore, this pipeline allows for efficient data synthesis, which, when combined with the data module’s conversion scripts, enables a closed-loop data workflow. 

\noindent \paragraph{Benchmark.} This component evaluates agent performance in \method environments. It supports a variety of tasks such as building, mining, and crafting, and offers both simple and challenging task modes to test agents under different levels of difficulty. The framework includes an automatic evaluation pipeline that leverages Vision-Language Models to analyze task videos, and provides batch task execution capabilities to run multiple tasks simultaneously and record completion videos. Additionally, it offers a quick benchmarking tool that simplifies the process of task execution and evaluation, enabling researchers to compare different agent models efficiently. 

\section{Comparison to Existing Interest of Minecraft}
\label{sec:comparison}

\begin{table}[ht]
\vspace{-0.5em}
\centering
\caption{Comparison of features across Minecraft development frameworks.}
\label{tab:framework_comparison}
\resizebox{\textwidth}{!}{%
\begin{tblr}{colspec={lcccc}, row{even}={bg=blue!5}, row{1}={font=\bfseries}}
\toprule
\textbf{Feature} & MineRL & MineDojo & Mineflayer & \textcolor{teal}{\method} \\
\midrule
\textbf{Observation/Action Space} & original & modified & modified & \textcolor{teal}{original} \\
\textbf{Dataset Features} & \makecell{annotated state-action pairs\\of human demonstrations}& \makecell{multimodal data\\scraped from Internet} & N/A & \makecell{\textcolor{teal}{efficient data structure}\\\textcolor{teal}{for both storing and loading}} \\
\textbf{Customizing Environments} & limited & easy & N/A & \textcolor{teal}{easy} \\
\textbf{Creating Agents} & difficult & difficult & easy & \textcolor{teal}{easy} \\
\textbf{Training Agents} & limited & limited & N/A & \makecell{\textcolor{teal}{pipelined, supporting both}\\ \textcolor{teal}{online and offline training}} \\
\textbf{Benchmarking Agents} & 11 tasks & 3000+ tasks & N/A & \textcolor{teal}{unlimited with MCU~\citep{mcu}} \\
\textbf{Evaluating Agents} & N/A & reward only & N/A & \makecell{\textcolor{teal}{customizable metrics with}\\ \textcolor{teal}{distributed framework}} \\
\textbf{Baseline Implementations} & N/A & N/A & N/A & \textcolor{teal}{supports a family of SOTA baselines} \\
\bottomrule
\end{tblr}%
}
\end{table}

To highlight the contributions of \method, we compare its features with those of other prominent Minecraft development frameworks, such as MineRL~\citep{minerl}, MineDojo~\citep{minedojo}, and Mineflayer~\citep{prismarinejs2013}, as shown in~\cref{tab:framework_comparison}. Although these frameworks offer valuable testbeds for Minecraft, they lack a seamless pipeline that integrates building, training, and evaluating agents, particularly when it comes to environment customization. Although \citet{vpt} extends MineRL by offering data loading and training code, its dataloader is primarily designed for demonstration purposes and is inefficient, while the RL training process remains closed-source and difficult to replicate. 
Moreover, the relatively slow speed of the MineRL and MineDojo simulators hampers RL training, adding considerable engineering challenges to Minecraft development. 
LLM-based agents~\citep{voyager,gitm} oversimplify the Minecraft problem by relying on Mineflayer APIs, whereas our goal is to build embodied agents that mirror the observation and control interface of human players. 
\section{Conclusions}
\label{sec:conclusion}

We introduced \method, a comprehensive and streamlined framework designed to advance the development of AI agents in Minecraft. By addressing the significant engineering challenges inherent in creating embodied policies for open-world environments, \method bridges the gap between conceptual algorithmic innovations and practical implementation. 

\newpage
\section*{Acknowledgments}

This work was supported by the National Science and Technology Major Project \textbf{\#\seqsplit{2022ZD0114902}} and the \textbf{CCF-Baidu Open Fund}. We sincerely appreciate their generous support, which enabled us to conduct this research.

\bibliography{refs}



\end{document}